%% file: _main.tex
\documentclass[11pt]{article}

\usepackage[final]{acl}

\usepackage[utf8]{inputenc} 
\usepackage[T1]{fontenc}    
\usepackage{hyperref}       
\usepackage{url}            
\usepackage{booktabs}       
\usepackage{times}
\usepackage{latexsym}
\usepackage{amsfonts}       
\usepackage{nicefrac}       
\usepackage{microtype}      
\usepackage{xcolor}         
\usepackage{multirow}
\usepackage{multicol}
\usepackage{graphicx}
\usepackage{colortbl}
\usepackage{wrapfig}
\usepackage{amsmath}
\usepackage{float}
\usepackage{makecell}
\usepackage{tcolorbox}
\usepackage{pifont}

\tcbuselibrary{skins, breakable}

\title{Enhancing Mathematical Problem Solving in LLMs through Execution-Driven Reasoning Augmentation}

\author{
  Aditya Basarkar \\
  North Carolina State University \\
  \texttt{avbasark@ncsu.edu}
  \And
  Benyamin Tabarsi \\
  North Carolina State University \\
  \texttt{btaghiz@ncsu.edu}
  \AND
  Tiffany Barnes \\
  North Carolina State University \\
  \texttt{tmbarnes@ncsu.edu}
  \And
  Dongkuan (DK) Xu \\
  North Carolina State University \\
  \texttt{dxu27@ncsu.edu}
}

\begin{document}

\maketitle

\input{sections/0_abstract}

\input{sections/1_introduction}
\input{sections/2_related_works}
\input{sections/3_IIPC}
\input{sections/4_experiments}
\input{sections/5_conclusion}
\input{sections/6_limitations}
\input{sections/9_ethics_statement}

\bibliography{references}

\newpage

\appendix

\input{appendix_sections/theoretical_overview}
\input{appendix_sections/extended_experiment_details}
\input{appendix_sections/additional_experimental_results}
\input{appendix_sections/case_studies}

\end{document}

%% file: sections/0_abstract.tex
\begin{abstract}

Mathematical problem solving is a fundamental benchmark for assessing the reasoning capabilities of artificial intelligence and a gateway to applications in education, science, and engineering where reliable symbolic reasoning is essential. Although recent advances in multi-agent LLM-based systems have enhanced their mathematical reasoning capabilities, they still lack a reliably revisable representation of the reasoning process. Existing agents either operate in rigid sequential pipelines that cannot correct earlier steps or rely on heuristic self-evaluation that can fail to identify and fix errors. In addition, programmatic context can distract language models and degrade accuracy. To address these gaps, we introduce Iteratively Improved Program Construction (\textit{IIPC}), a reasoning method that iteratively refines programmatic reasoning chains and combines execution feedback with the native Chain-of-thought abilities of the base LLM to maintain high-level contextual focus. IIPC surpasses competing approaches in the majority of reasoning benchmarks on multiple base LLMs. All code and implementations are released as \href{https://github.com/ncsu-dk-lab/IIPC-Math-Reasoning-Agent}{open source}.

\end{abstract}

%% file: sections/1_introduction.tex
\section{Introduction}

Mathematical reasoning, the ability to reliably solve multi-step math problems through symbolic manipulation and language-based deduction, is a necessary skill for LLM-based systems to master in order to build generally intelligent systems that can automate intellectual tasks. It is especially crucial in applications such as scientific discovery~\cite{cherian2024llmphycomplexphysicalreasoning, shojaee2025llmsrscientificequationdiscovery}, optimization~\cite{huang2024largelanguagemodelmeets, forootani2025surveymathematicalreasoningoptimization, met13030490}, financial modeling~\cite{yao2025evaluatinglargelanguagemodels, santos2022portoptim}, and education~\cite{gupta2025finalanswersevaluatinglarge, liu2025sizedoesntfitall}, among many others. As research advances towards enabling language models to use tools and understand abstract concepts, mathematical reasoning emerges as a benchmark for assessing their ability to combine symbolic structure with logical deduction. Advancing this capability is necessary for building AI systems that can engage in effective, applicable, and scientifically grounded reasoning.

Despite substantial progress, many reasoning agents still struggle with two limitations. 
First, most systems lack a \textbf{persistent and reliably manipulable representation of their overall reasoning state} that allows them to make informed revisions of earlier steps. Once a model commits to a reasoning chain, whether expressed as text, a sequence of agentic steps, or a program, subsequent revisions are typically superficial, based on heuristic self-evaluation or nonexistent. Without an editable global reasoning state, the model cannot identify or act on contradictions, invalid computations, or revise previous steps, making it vulnerable to cascading errors~\cite{cemri2025multi}. Second, execution-guided agents lack \textbf{stabilizers against program bias, often over-prioritizing execution signals} that could be logically flawed over token-level reasoning, resulting in brittle reasoning trajectories. When program outputs are used to draw conclusions, irrelevant or incorrect information may cause biased reasoning in models, reducing accuracy~\cite{shi2023irrelcontext, yang2025llmdistract}.

Existing approaches address portions of these limitations, but leave important gaps. Multi-agent systems such as Cumulative Reasoning (CR)~\cite{zhang2023cumulative} and Multi-agent Condition Mining (MACM)~\cite{lei2024macm} have enabled stepwise deliberation. However, their sequential structure locks their reasoning trajectory into a fixed forward direction. This rigidity prevents agents from freely revising earlier steps, making them vulnerable to cascading errors, a phenomenon identified by~\cite{cemri2025multi} as "inter-agent misalignment". Agents like Self-Refine~\cite{madaan2023selfrefine}, Reflexion~\cite{shinn2023reflexionlanguageagentsverbal}, Tree-of-Thoughts~\cite{yao2023tree}, and Graph-of-Thoughts~\cite{besta2024graph} offer self-evaluation, refinement, and backtracking capabilities, but their effectiveness is bounded by their capacity to produce accurate self-evaluations~\cite{huang2023large}, often resulting in unreliable refinements and trajectory selections. Program and tool-based agents such as Program of Thoughts (PoT)~\cite{chen2023pot}, Tool Integrated Reasoning Agent (ToRA)~\cite{gou2024tora}, Code-based Self Verification (CSV)~\cite{zhou2023codeselfver}, and Program-aided Language Model (PAL)~\cite{gao2023palprogramaidedlanguagemodels} provide deterministic signals through code execution. These programs are usually one-off artifacts and are not subjected to targeted revisions, leading to rigidity similar to that of sequential reasoning agents. Furthermore, recent studies have demonstrated that such agents, particularly those relying on program-based reasoning, are susceptible to irrelevant or misleading context, often over-incorporating flawed information into their reasoning traces and causing performance degradation~\cite{shi2023irrelcontext, yang2025llmdistract}.

To overcome these limitations, we propose Iteratively Improved Program Construction (IIPC), an agent that treats programs as explicit representations of the model's reasoning chain. While most agents generate one-off code blocks, IIPC generates reasoning chains as programs designed for transparent inspection and evolution through informed revision. Each revision is grounded in deterministic feedback that includes information about intermediate steps, allowing the model to correct errors or inconsistencies in its reasoning and make causally informed changes. A memory of past mistakes is maintained to ensure that refinements minimize "revisit regret" and steer improvements away from failure modes rather than resampling failed programs. To preserve a high-level contextual focus, IIPC adopts a dual-branch architecture. The token-level reasoning branch produces a CoT reasoning trace free from dependence on program output, while the program-refinement branch iteratively refines an executable representation of the model's reasoning. The outputs of the two branches are combined only at the final stage to provide context for the final reasoning chain. Through this mechanism, IIPC avoids over-reliance on potentially incorrect or irrelevant program results while leveraging relevant information. As a result, IIPC provides a unified reasoning mechanism that combines manipulable representations of reasoning traces with context-stable reasoning, overcoming the limitations of existing multi-agent, self-evaluating, and execution-guided systems.

The paper's primary contributions are as follows:
\begin{enumerate}
    \item We introduce IIPC, a new reasoning method that is designed to refine programs through execution-guided feedback, integrate execution outputs into its own reasoning abilities, and surpass other code-based, state-of-the-art, non-ensemble reasoning agents in difficult math problem-solving benchmarks.
    \item We provide the reasoning-trace corpus generated from testing IIPC, including problem statements, initial propositions, generated code, execution outputs, integrated deliberation, and final answers, to facilitate reproducible evaluation, detailed error analysis, and future work on program-centric reasoning.
    \item We conduct a comprehensive evaluation of IIPC against various multi-agent reasoning methods across large language models and mathematical reasoning benchmarks. IIPC surpasses PoT, MACM, and CR on the majority of the mathematical problem-solving benchmarks on multiple base LLMs.
\end{enumerate}

%% file: sections/2_related_works.tex
\section{Related Work}
Recently developed multi-agent systems vary in their approach to structuring mathematical reasoning. CR~\cite{zhang2023cumulative} uses three components: a proposer, verifier, and reporter. These components generate, evaluate, and collect reasoning steps. CR achieves 72.2\% accuracy on 500 MATH problems~\cite{hendrycksmath2021} using GPT-4 and early LLaMA models. MACM~\cite{lei2024macm} accumulates and verifies conditions needed to solve the given problem rather than processing the reasoning steps. MACM also uses voting mechanisms to leverage uncertainty and explore multiple reasoning trajectories that are higher in quality than CoT~\cite{wei2022chain}, Tree-of-Thoughts (ToT)~\cite{yao2023tree}, or Graph-of-Thoughts (GoT)~\cite{besta2024graph} prompting. Despite their improvements on popular benchmarks, sequential processing introduces rigid reasoning chains that cause errors to propagate over subsequent steps.

While sequential multi-agent systems like MACM and CR focus on reasoning through iterative step accumulation, other research directions enhance problem-solving by integrating external tools. Tool-Integrated Reasoning Agent (ToRA)~\cite{gou2024tora} fine-tunes LLaMA models to interleave natural language reasoning with tool calls. PoT~\cite{chen2023pot} generates and executes Python programs, using the outputs to arrive at the final answer. Code-Based Self-Verification (CSV)~\cite{zhou2023codeselfver} includes programmatic checks to verify and refine solutions, improving reliability through iterative code-driven validation.

Other approaches aim to improve reasoning internally through self-correction. Agents like self-refine~\cite{madaan2023selfrefine} and Reflexion~\cite{shinn2023reflexionlanguageagentsverbal} adopt an iterative self-improvement approach, generating feedback on their own output and refining it until a predefined stopping condition is met. These methods provide foundational insights for enhancing program-based reasoning, which our approach leverages and advances.

IIPC's program refinement mechanism is similar to CodePRM~\cite{li-etal-2025-codeprm} in its usage of execution feedback to refine reasoning. However, IIPC treats programs as representations of the reasoning chain, whereas CodePRM edits natural-language thoughts based on their quality scores for better downstream execution. The dual-branch architecture used by IIPC conceptually resembles the Talker-Reasoner architecture~\cite{chris2024talkreason}, where a talker agent performs fast language generation, the reasoner performs multi-step reasoning, and they interact through a shared memory. However, unlike the interleaved process of the Talker-Reasoner, IIPC maintains a clean CoT trace and only merges at the end to remain robust to noisy programmatic outputs.

Execution-Guided Classifier-Free Guidance (EG-CFG)~\cite{egcfg_lavon} is a decoding strategy that guides token sampling based on line-level execution. OpenCodeInterpreter (OCI)~\cite{oci_zheng} uses an execution guided refinement loop to iteratively refine generated programs. Outcome-Refining Process Supervision (ORPS)~\cite{orps_yu} performs a beam search over reasoning-code trajectories, selecting candidates based on execution-based outcome signals. In contrast, IIPC treats programs as representations for mathematical problem solving, refining them using both process- and outcome-level signals to identify and correct reasoning errors.

%% file: sections/3_IIPC.tex
\section{Iteratively Improved Program Construction (IIPC)}

\begin{figure*}[t]
  \centering
  \includegraphics[width=0.65\textwidth,height=0.45\textwidth]{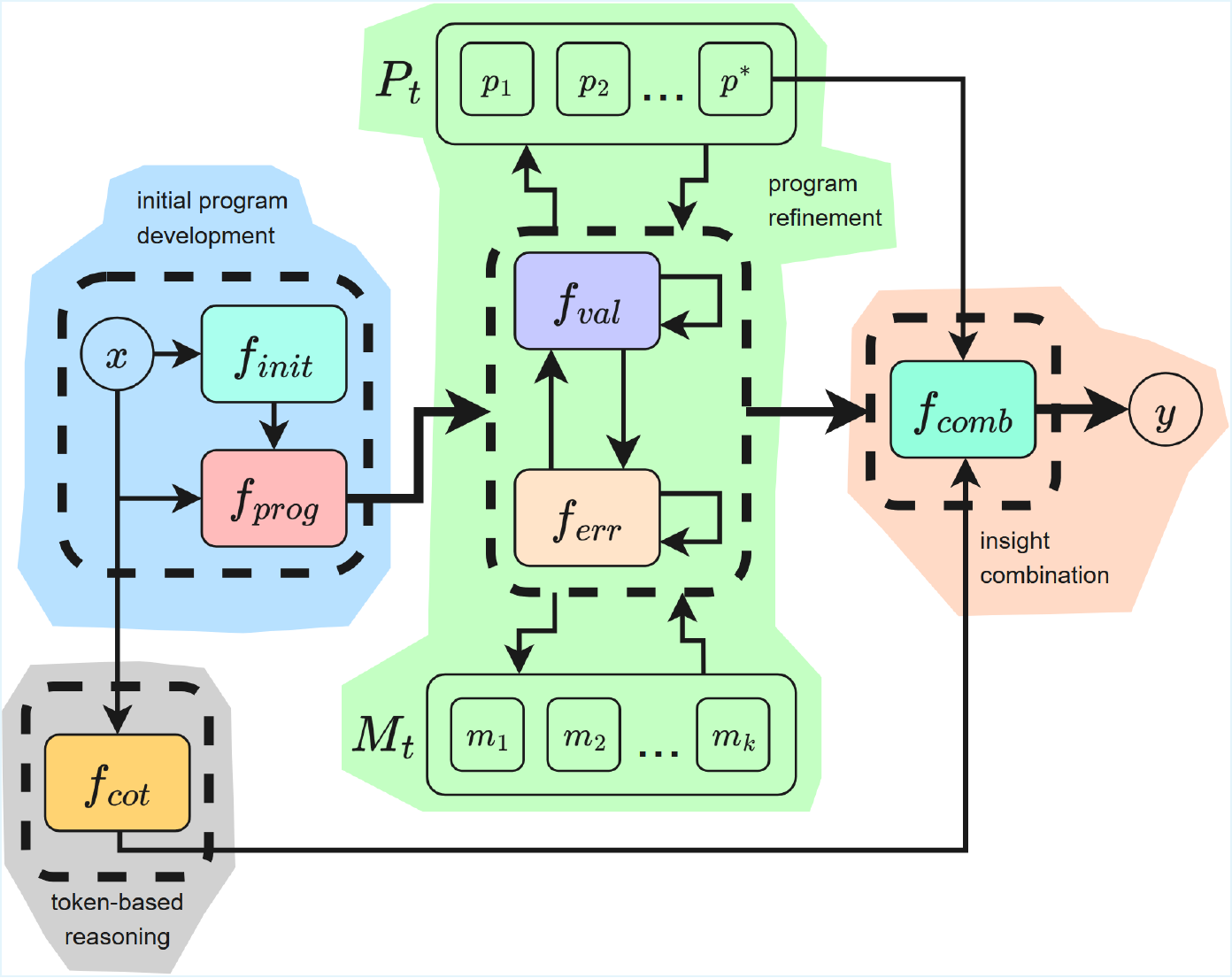}
  \caption{
    Overview of IIPC. 
    $f_{\text{init}}$ derives key propositions from the problem statement; $f_{\text{prog}}$ generates an initial candidate program; $f_{\text{val}}$ evaluates program correctness and logical consistency; if errors are detected, the error correction component $f_{\text{err}}$ revises the program accordingly; $f_{\text{cot}}$ produces a textual chain of thought;$f_{\text{comb}}$ combines program and token reasoning context for final output; $M_t$ denotes the error descriptor memory at refinement step $t$; $P_t$ represents the program store at step $t$.
    }
  \label{fig:iipc-flow}
\end{figure*}

Our proposed method, IIPC, introduces an execution-guided reasoning methodology that directly addresses the limitations of works previously mentioned. Rather than accumulating steps or conditions sequentially, as in CR and MACM, IIPC develops programs that solve the given problem while simultaneously storing compact failure descriptors. This allows IIPC to iteratively converge toward correct solutions while dynamically avoiding previously failed reasoning paths.

IIPC begins by using a model $f$ for developing a set of initial propositions $s$ ($s = f_{init}(x)$), which are statements derived explicitly from the given problem $x$, as well as additional information needed to solve the problem. These statements encapsulate all critical information, whether explicitly provided by the problem or implicitly required, to guide subsequent solution steps.

Given the problem statement and the set of initial propositions, IIPC generates a candidate solution program $p_1$ ($p_1 = f_{prog}(x, s)$) that transparently represents its reasoning process. Every generated program conforms to the following constraints:

\begin{itemize}
  \item Programs must be restricted to the following libraries: numpy, math, sympy, scipy, and scikit-spatial. These libraries are sufficient to express the vast majority of mathematical reasoning tasks. Additionally, they are stable, widely used, and deterministic, minimizing variability across runs.
  \item To facilitate debugging, the program must avoid using any list comprehensions or recursion and should make use of print statements.
  \item Comments must be verbose and descriptive for aiding correction and refinement.
  \item The program must include a section that verifies the final answer, allowing the framework to evaluate its correctness.
\end{itemize}

Once the program is constructed, it is executed using an interpreter $E$ to yield an output ($o_1 = E(p_1)$), which can either be the program's output, an error, or both. This serves as the basis for subsequent reflection and refinement.

IIPC reviews the generated program’s reasoning and consults a persistent reflection memory containing past mistakes ($M_t$, where $ t$ is the time point of the current refinement step) to avoid repeating unsuccessful strategies. It then either (1) issues a revised program accompanied by a concise reflection on the identified flaw, which is logged in the memory store before the revised program is re-executed, or (2) outputs a stopping signal indicating that the current program and its outputs are satisfactory for solving the problem without further refinement. If the program produces an error, IIPC amends only the offending code segment without modifying the reasoning chain itself. It also produces a concise reflection on the mistake, which is appended to its memory store for guiding future iterations.

IIPC iteratively refines the design by executing a series of process validations and error corrections to produce the optimal program for solving the given problem. If a program does not result in an error, the \textit{process validation} component will assess the program's validity and its output. However, if the program produces an error, the \textit{error correction} component will revise it while preserving the reasoning chain. The resulting programs from either process are rerun and cycled (along with their output and error) back into one of the two components for further correction and/or enhancement. The current implementation for refinement, as tested in the main experiments, allows a maximum of two process validations and two error corrections after each process validation. All working programs and their output are stored in an array $P_t$, with the program with the highest index being the most recently developed working program. This back-and-forth loop between the process validation and error-correction components continues until a stop condition is met. The \textit{stop condition} could either be a satisfactory program, a limit on the allowed number of refinements or error corrections, or a limit on the token usage of the agent. Formally, this refinement process can be summarized in the following formulas:
\begin{align}
  (p_{t+1}, m_{t+1}) &=
  \begin{cases}
    f_{err}(x, s, p_t, o_t, M_t), & \text{if } o_t \in e, \\[4pt]
    f_{val}(x, s, p_t, o_t, M_t), & \text{if } o_t \notin e
  \end{cases} \\
  M_{t+1} &= M_t \cup m_{t+1} \\[4pt]
  P_{t+1} &= P_t \cup p_t \ \ \text{if} \ \ o_t \notin e
\end{align}

\vspace{0.5em}

where $m_{t+1}$ is an error or mistake descriptor, $f_{err}$ is the function that corrects errors, and $f_{ref}$ is the refinement function that can result in a $\emptyset$ for both $p_{t+1}$ and $m_{t+1}$ if the refinement process reaches a stopping point. In addition to program construction and refinement, IIPC generates a purely text-based CoT $c$ in a separate branch and arrives at a temporary answer ($c = f_{cot}(x, s)$). The most recent working program ($p^\star \in P_t$) and its output ($o^\star$) with the generated CoT ($c$) text is concatenated through a structured integration prompt, which is used as input to the underlying LLM to arrive at the final answer for the problem ($y = f_{comb}(x,s,p^\star,o^\star,c)$). This step allows IIPC to reevaluate the solution using symbolic evidence from both refined program execution and native linguistic reasoning, thereby reducing over-dependence on a single kind of context.

%% file: sections/4_experiments.tex
\section{Experiments}

\paragraph{Experimental Setup.} Our evaluation benchmarks the IIPC framework on two mathematical reasoning datasets: (1) MATH~\cite{hendrycksmath2021}, comprising complex problems in multiple subjects with five difficulty levels; (2) and AIME (American Invitational Mathematics Examination)~\cite{veeraboina2024aime}, consisting of difficult competition math problems from 1983 to 2024 that emphasize creative problem solving and is more challenging than the MATH dataset. For the MATH dataset, we evaluate on a balanced subset of the official test split from the original dataset, targeting uniform coverage across all topic-difficulty combinations (35 bins total). Due to limited availability in certain bins, the final subset contained 1483 problems total. For the AIME dataset~\cite{veeraboina2024aime}, we use the complete set of 933 problems spanning the years 1983-2024. We compare IIPC's performance against other reasoning paradigms, including CR~\cite{zhang2023cumulative}, MACM~\cite{lei2024macm}, and PoT prompting~\cite{chen2023pot}. We evaluate across multiple state-of-the-art LLMs, including GPT-4o mini, Gemini 2.0 Flash, Mistral Small 3.2 24B, Gemma 3 27B, and Llama 4 Maverick, to characterize how model architecture influences accuracy. Finally, our ablations isolate the contributions of specific architectural components, examine the effect of varying decoding temperatures, evaluate all methods under voting-based aggregation, and use GSM8K~\cite{cobbe2021gsm8k}, a data set consisting of grade school math problems, to assess performance degradation due to agentic overhead on simpler math problems.

\paragraph{Agent Implementations.} For the purposes of this study, the main evaluation tests all prompting-based reasoning agents with access to code interpreters, excluding multi-trajectory aggregation (voting) mechanisms due to their high token usage and costs. To comply with this constraint, reasoning agents like MACM~\cite{lei2024macm}, which utilize and aggregate multiple answers, are adapted to operate within a single run and evaluated accordingly. While this may handicap the MACM agent, which is designed to operate in an ensemble environment, testing its performance in these conditions allows for a fair comparison with other such agents and provides insights into the weaknesses of processing steps sequentially rather than in parallel. As mentioned before, a separate ablation is provided to assess the effects of voting environments on the performance of all agents. Our implementation of PoT includes an error-correction loop that retries program generation when execution fails, until it reaches a fixed retry limit and has to fall back to standard Chain-of-Thought reasoning. This design allows us to better compare the mathematical reasoning capabilities of agents against PoT while controlling for coding ability. Our custom implementations of MACM and CR attempt to incorporate as much of the original implementations as possible and follow the proposed methods closely (with the exception of the voting mechanism for MACM). All agents are tested with an LLM decoding temperature of 0.1 and measure performance as accuracy (the number of correctly solved questions divided by the total number of questions evaluated).

\paragraph{Assessing Correctness.} To assess answer correctness, we first use deterministic equivalence and fall back to LLM-based judging using the LLaMA 4 Maverick model if necessary. Although deterministic evaluation is ideal, code-centric agents may produce equivalent answers in differing formats (e.g., 1.4142 versus $\sqrt{2}$). To ensure transparency and reproducibility, we fix the decoding temperature at 0 for the LLM judge and release the full judging code. On the AIME dataset, 97.78\% of answers were verified via exact integer matching. On the MATH dataset, 72.85\% of the generated answers were deterministically verified; LLM judging was only used to confirm whether the answers were incorrect after deterministic judging failed. On GSM8K, 89.35\% of the answers were verified using exact answer matching.

\paragraph{Other Considerations.} While our evaluations focus on accuracy and tend to favor reasoning stability over token efficiency, we do recognize the fundamental limitation that IIPC incurs significant token costs by regenerating refined or corrected programs with every iteration. Additionally, our evaluation reports single-run accuracy on fixed benchmarks. However, while gains may be small for some evaluations, IIPC consistently outperforms other agents across all LLMs and datasets, reducing the likelihood that performance gains are due to noise.

\definecolor{lightgray}{gray}{0.9}
\definecolor{teal}{RGB}{0,128,128}

\definecolor{lightgray}{gray}{0.9}
\definecolor{teal}{RGB}{0,128,128}

\begin{table*}[t]
  \centering
  \caption{Accuracy (\%) of reasoning methods across five LLMs. 
  IIPC row is shaded; bold teal = per-column SOTA. 
  Numbers in parentheses show $\Delta$ vs PoT baseline.}
  \label{tab:stacked-results}
  \setlength{\tabcolsep}{12pt}
  \small

  \textbf{MATH} \\[0.25em]
  \begin{tabular}{@{}lccccc@{}}
    \toprule
    \makecell{\textbf{Reasoning}\\\textbf{System}} &
    GPT-4o-mini & Gemini 2.0 Flash & Mistral 3.2 24B & Gemma 3 27B & Llama 4 Maverick \\
    \midrule
    PoT   & \textbf{\textcolor{teal}{81.19}} & 92.58 & 89.62 & 89.01 & 88.94 \\
    CR    & 76.53 (-4.66) & 90.09 (-2.49) & 83.61 (-6.01) & 87.05 (-1.96) & 89.94 (+1.00) \\
    MACM  & 72.62 (-8.57) & 90.09 (-2.49) & 82.13 (-7.49) & 86.72 (-2.29) & 88.67 (-0.27) \\
    \rowcolor{lightgray}
    IIPC  & 80.98 (-0.21) & \textbf{\textcolor{teal}{94.13}} (+1.55) & \textbf{\textcolor{teal}{90.83}} (+1.21) & \textbf{\textcolor{teal}{90.56}} (+1.55) & \textbf{\textcolor{teal}{91.23}} (+2.29) \\
    \bottomrule
  \end{tabular}

  \vspace{1em}

  \textbf{AIME} \\[0.25em]
  \begin{tabular}{@{}lccccc@{}}
    \toprule
    \makecell{\textbf{Reasoning}\\\textbf{System}} &
    GPT-4o-mini & Gemini 2.0 Flash & Mistral 3.2 24B & Gemma 3 27B & Llama 4 Maverick \\
    \midrule
    PoT   & \textbf{\textcolor{teal}{31.40}} & 59.16 & 48.12 & 46.20 & 62.49 \\
    CR    & 23.90 (-7.50) & 53.48 (-5.68) & 40.09 (-8.03) & 41.69 (-4.51) & 62.17 (-0.32) \\
    MACM  & 17.79 (-13.61) & 51.98 (-7.18) & 37.62 (-10.50) & 41.69 (-4.51) & 62.17 (-0.32) \\
    \rowcolor{lightgray}
    IIPC  & 29.05 (-2.35) & \textbf{\textcolor{teal}{64.20}} (+5.04) & \textbf{\textcolor{teal}{52.52}} (+4.40) & \textbf{\textcolor{teal}{50.48}} (+4.28) & \textbf{\textcolor{teal}{69.77}} (+7.28) \\
    \bottomrule
  \end{tabular}
\end{table*}

\begin{figure}[t]
  \centering
  \includegraphics[width=\columnwidth]{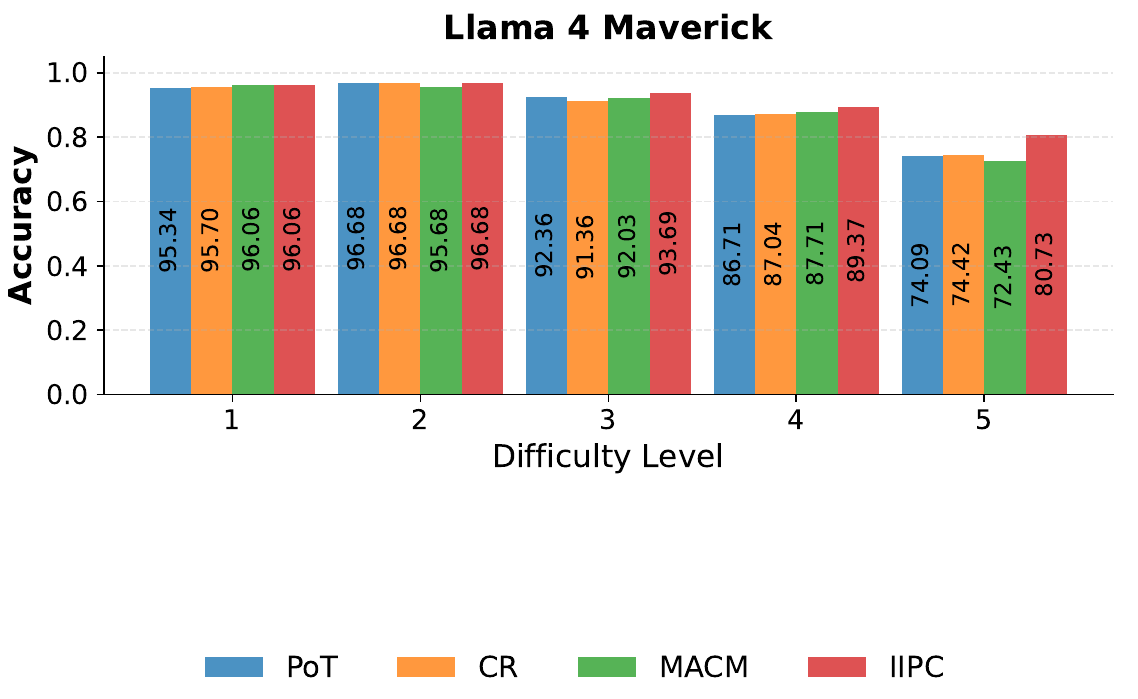}
  \caption{Accuracy of PoT, IIPC, CR, and MACM on the MATH dataset using Llama 4 Maverick. This bargraph is stratified by difficulty level}
  \label{fig:llama4mav_bar}
\end{figure}

\begin{figure}[t]
  \centering
  \includegraphics[width=0.8\columnwidth]{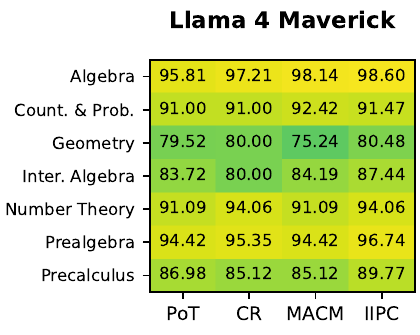}
  \caption{Heatmap of accuracy (\%) by subject area on the MATH benchmark for Llama-4-Maverick. Columns correspond to reasoning agents (PoT, IIPC, CR, and MACM) and rows correspond to mathematical domains.}
  \label{fig:llama4mav_heatmap}
\end{figure}

\subsection{MATH}

The MATH benchmark includes problems across a broad spectrum of advanced topics, including number theory, geometry, algebra (from prealgebra through intermediate algebra), precalculus, and counting \& probability, each presented at five increasing difficulty tiers. On this benchmark, IIPC outperformed the other reasoning methods across most LLM backbones, achieving the highest scores on Gemini 2.0 Flash (94.13\%), Mistral Small 3.2 24B (90.83\%), Gemma 3 27B (90.56\%), and Llama 4 Maverick (91.23\%). Conversely, PoT surpassed IIPC and all other methods on GPT-4o-mini (81.19\%). This suggests that, in models with lower reasoning capacity, the additional complexity introduced by IIPC can become counterproductive, whereas PoT reasoning traces appear better aligned with these models' capabilities, providing just enough scaffolding without overwhelming their reasoning bandwidth.

To further analyze agent behavior on the MATH benchmark, we break down performance by difficulty level and topic. Although the main evaluation presents this analysis only for Llama 4 Maverick, analogous results for all other language models are provided in the appendix. As evident in figure~\ref{fig:llama4mav_bar}, IIPC consistently matches or outperforms other agents as difficulty increases. On the most challenging level-5 problems, IIPC achieves an accuracy of 80.73, outperforming the next best method by 6.31 points. As shown in figure~\ref{fig:llama4mav_heatmap}, we find that with the exception of Counting and Probability, IIPC achieves the highest accuracy across all domains. Most notably, we record the highest gains in Pre-calculus (+2.79\%, 89.77\%) and Intermediate Algebra (+3.25\%, 87.44\%). This provides further evidence for IIPC's ability to solve difficult problems and its effectiveness over existing methods.

\subsection{AIME}

The AIME dataset contains problems from the American Invitational Mathematics Examination from 1983 to 2024. It emphasizes creative problem-solving and is the most difficult of the 3 datasets tested. Consistent with its performance on the MATH benchmark, IIPC achieved the highest accuracy on most models, specifically Gemini 2.0 Flash (64.20\%), Mistral 3.2 24B (52.52\%), Gemma 3 27B (50.48\%), and Llama 4 Maverick (69.77\%). In contrast, PoT surpassed all other methods on GPT-4o-mini (31.40\%). Similar to the results on the MATH benchmark, GPT-4o-mini may lack the reasoning capacity required to fully benefit from IIPC’s more refined loop. IIPC’s design scales well to highly challenging mathematical reasoning tasks when supported by sufficiently capable models, while PoT seems to remain more robust on models with more constrained reasoning capacity.

\subsection{Ablation Tests}

\begin{table}[t]
  \centering
  \caption{Accuracy (\%) of reasoning methods on GSM8K. 
  Bold teal = per-row SOTA. Numbers in parentheses show $\Delta$ vs CoT baseline.}
  \label{tab:iipc-cot-ablation-singlecol}
  \setlength{\tabcolsep}{5pt}
  \renewcommand{\arraystretch}{1.05}
  \small

  \begin{tabular}{@{}lccccc@{}}
    \toprule
    \textbf{LLM} & \textbf{CoT} & \textbf{IIPC} & \textbf{PoT} & \textbf{MACM} & \textbf{CR} \\
    \midrule
    GPT-4o-mini      & \textbf{\textcolor{teal}{94.54}} & 94.24 & 93.56 & 91.05 & 92.34 \\
    Gemini 2.0 Flash & \textbf{\textcolor{teal}{96.06}} & 95.60 & 95.75 & 92.57 & 91.89 \\
    Mistral 3.2 24B  & \textbf{\textcolor{teal}{95.22}} & 95.00 & 94.84 & 93.86 & 94.47 \\
    Gemma 3 27B      & \textbf{\textcolor{teal}{95.45}} & 94.84 & 94.47 & 91.36 & 93.18 \\
    Llama 4 Maverick & \textbf{\textcolor{teal}{96.13}} & 95.53 & 95.53 & 94.24 & 95.91 \\
    \bottomrule
  \end{tabular}
\end{table}

\begin{table}[t]
  \centering
  \caption{Ablation results on MATH (Gemini 2.0 Flash) and AIME (Gemini 2.0 Flash, GPT\mbox{-}4o\mbox{-}mini). Best per column is \textbf{\textcolor{teal}{bold teal}}.}
  \label{tab:iipc-ablation}
  \setlength{\tabcolsep}{5pt}
  \small
  \resizebox{\linewidth}{!}{%
  \begin{tabular}{@{}lccc@{}}
    \toprule
    \textbf{Methods} & \textbf{MATH} (Gemini) & \multicolumn{2}{c}{\textbf{AIME}} \\
    \cmidrule(lr){2-2}\cmidrule(lr){3-4}
                     & Gemini 2.0 Flash & Gemini 2.0 Flash & GPT-4o-mini \\
    \midrule
    CoT         & 92.92 & 56.70 & 21.11 \\
    PoT-NC      & 92.52 & 59.12 & 31.19 \\
    PoT         & 92.58 & 59.16 & \textbf{\textcolor{teal}{31.40}} \\
    IIPC-NS-NMS & 93.19 (+0.61) & 60.77 (+1.61) & 26.90 (-4.50) \\
    IIPC-NS     & 93.59 (+1.01) & 61.52 (+2.36) & 27.12 (-4.28) \\
    IIPC        & \textbf{\textcolor{teal}{94.13}} (+1.55) & \textbf{\textcolor{teal}{64.20}} (+5.04) & 29.05 (-2.35) \\
    \bottomrule
  \end{tabular}}
\end{table}

\begin{table}[t]
  \centering
  \caption{IIPC temperature ablation on AIME (Gemini 2.0 Flash). Best is \textbf{\textcolor{teal}{bold teal}}.}
  \label{tab:temperature-ablation}
  \setlength{\tabcolsep}{6pt}
  \small
  \begin{tabular}{@{}lc@{}}
    \toprule
    \textbf{Temperature} & \textbf{Accuracy (\%)} \\
    \midrule
    0.1 & 64.20 \\
    0.3 & \textbf{\textcolor{teal}{64.52}} \\
    0.5 & 63.56 \\
    0.7 & 63.13 \\
    0.9 & 63.24 \\
    \bottomrule
  \end{tabular}
\end{table}

\begin{table}[t]
  \centering
  \caption{Effect of voting on MATH Level-5 (245 problems) using Llama 4 Maverick. Best in each column is \textbf{\textcolor{teal}{bold teal}}.}
  \label{tab:voting-ablation}
  \setlength{\tabcolsep}{6pt}
  \small
  \begin{tabular}{@{}lcc@{}}
    \toprule
    \textbf{Agent} & \textbf{No Voting} & \textbf{Voting} \\
    \midrule
    CoT  & 70.61 & 74.29 \\
    PoT  & 73.47 & 80.30 \\
    CR   & 68.57 & 78.37 \\
    MACM & 69.80 & 76.33 \\
    IIPC & 78.78 & 80.30 \\
    \bottomrule
  \end{tabular}
\end{table}

To better understand which components of IIPC contribute most to its performance gains, we conducted a series of ablation studies on both the MATH and AIME benchmarks. The first set of ablations incrementally adds architectural features across five methods:

\begin{itemize}
    \item CoT: A baseline Chain-of-Thought method with purely token-level reasoning.
    \item PoT-NC: A Program-of-Thoughts method that generates and executes programs but omits the correction loop.
    \item PoT: A PoT method with both program generation and execution-guided error correction.
    \item IIPC-NS-NMS: An IIPC variant with the full correction and iterative refinement loop but without the token-reasoning branch or persistent reflection memory. Conclusions are made directly from the programmatic context.
    \item IIPC-NS: An IIPC variant with the full correction and iterative refinement loop without the token-reasoning branch, but with the persistent reflection store. Conclusions are made directly from the programmatic context.
    \item IIPC: The full IIPC reasoning agent.
\end{itemize}

\paragraph{Architectural Ablation Analysis.} Table~\ref{tab:iipc-ablation} summarizes the impact of key IIPC mechanisms across datasets and model backbones. On Gemini 2.0 Flash, the iterative refinement, reflection memory, and dual-branch separation provide substantial gains. Iterative refinement improves accuracy by +0.61\% on the MATH dataset (92.58 → 93.19) and +1.61\% on the AIME dataset (59.16 → 60.77). The reflection memory improves accuracy by an additional 0.4\% on the MATH dataset (93.19 → 93.59) and 0.75\% on the AIME dataset (60.77 → 61.52). Dual-branch separation yields an additional +0.54\% on the MATH dataset (93.59 → 94.13) and +2.68\% on the AIME dataset (61.52 → 64.20). Together, these mechanisms increase performance by 1.55\% over PoT on the MATH dataset (92.58 → 94.13),  and by 5.04\% over PoT on the AIME dataset (59.16 → 64.20), highlighting their effectiveness for tasks requiring structured reasoning. On GPT-4o mini, while PoT achieves the highest accuracy, ablations show that individual IIPC components partially mitigate the performance gap. The reflection store and dual-branch separation reduce the accuracy drop of IIPC relative to PoT, even though the accuracy of the full IIPC agent remains below the PoT baseline.

\paragraph{Complexity Ablation Analysis.} We also compared IIPC, MACM, CR, and PoT against CoT on GSM8K, as shown in Table~\ref{tab:iipc-cot-ablation-singlecol}, to determine whether agent complexity hinders accuracy compared to chain of thought reasoning on easier math problems. CoT performs near the upper limit across most models, achieving 95–97\% accuracy on the strongest backbones. On average, IIPC and PoT closely trail CoT's performance, while MACM and CR have slightly lower accuracy. These results suggest that GSM8K’s problem structure saturates the benefits of iterative refinement, leaving little room for improvement beyond direct token-level reasoning on stronger models. Nonetheless, IIPC’s close performance to CoT demonstrates that its structure does not significantly degrade accuracy, even in low-complexity reasoning environments.

\paragraph{Temperature Ablation Analysis.} We further examined the effect of varying the decoding temperature on IIPC’s performance with Gemini 2.0 Flash on the AIME benchmark (Table~\ref{tab:temperature-ablation}). Across the tested range (0.1–0.9), accuracy remained relatively stable, fluctuating within a narrow band of 1.39 percentage points. The best performance was achieved at 0.3 (64.52\%), while higher temperatures (0.5–0.9) yielded slightly lower accuracies (63.13\%–63.56\%). These results suggest that IIPC is robust to temperature changes, with only marginal sensitivity, although higher temperatures appear to slightly diminish performance.

\paragraph{Voting Ablation Analysis.} Beyond architectural and temperature-based variations, we also evaluate the effect of voting, a method used in the original MACM implementation, on IIPC’s performance. As detailed in the appendix, our evaluation excludes stochastic multi-trajectory aggregation due to high costs, but we isolate its effect in a smaller ablation study using Llama-4-Maverick on 245 Level-5 MATH problems (35 from each topic). To align with MACM’s original setting, we use a decoding temperature of 0.1 for the non-voting environment, 0.7 for the voting environment, a minimum of 5 voters, and a maximum of 7 voters for tie-breaking. The results in Table~\ref{tab:voting-ablation} show that without voting, IIPC achieves the highest accuracy at 78.78\%, outperforming even voting-based variants of many baselines. When voting is enabled, IIPC achieves the same accuracy as PoT at 80.30\%. Voting yields larger gains for PoT (+6.83\%), CoT (+3.68\%), CR (+9.8\%), and MACM (+6.53\%), while the relative improvement for IIPC (+1.52\%) is smaller. IIPC exhibits smaller gains from voting, suggesting reduced single-trajectory variance compared to baselines that rely more on sampling diversity.

%% file: sections/5_conclusion.tex
\section{Conclusion}

In this work, we introduced Iteratively Improved Program Construction (IIPC) for mathematical reasoning with LLMs through execution-guided refinement while mitigating execution bias by reducing over-dependence on program-based context. Our experiments show that IIPC consistently delivers competitive or superior performance. On GSM8K, where simpler reasoning tasks leave limited room for improvement, IIPC demonstrated that its additional structure does not significantly hinder its reasoning capabilities compared to simpler baselines. On both the MATH and AIME datasets, IIPC achieved the highest accuracy across 4 of the 5 models tested. These results demonstrate that integrating token-based reasoning with iterative program refinement provides LLMs with valuable capacity for trajectory correction and reasoning-level regularization, reducing over-commitment to execution-conditioned context.

%% file: sections/6_limitations.tex
\section{Limitations}
Despite the advantages of IIPC, our results also revealed some trade-offs. On GPT-4o mini, IIPC lagged behind PoT on both MATH and AIME. These findings suggest that while IIPC scales effectively with models capable of sustaining its iterative refinement loop, models with limited reasoning or coding abilities may instead benefit from PoT’s simpler structure. In addition, although IIPC's design enables advanced refinement for solving difficult problems, it is a token-intensive approach to reasoning.
While IIPC proves robust on GSM8K and state-of-the-art on MATH and AIME across several high-capacity models, there is room for improvement in subsequent research. Future directions include addressing token efficiency in IIPC's design, adapting to models’ reasoning capacity, and extending IIPC to other domains requiring verifiable, structured reasoning beyond mathematics.

%% file: sections/9_ethics_statement.tex
\section{Ethical Considerations}

Our work introduces an execution-guided agent for enhanced mathematical reasoning in LLMs. The proposed method is evaluated on publicly available benchmarks (AIME, MATH, GSM8K) and does not involve human subjects, personal data, or proprietary datasets. We have ensured that none of the data used or generated by this research contains information that uniquely identifies people. While IIPC is designed to improve reasoning capabilities, it doesn't guarantee correctness. Any outputs generated by IIPC should not be relied on without independent verification, especially in high-stakes or real-world decision-making contexts. 

ChatGPT was used to provide feedback on quality, assist with some grammatical refinement and articulation, and support understanding and exploration of existing mathematical concepts for framing the proposed method. However, all revisions were carefully considered and manually done. All ideas, experimental designs, analyses, and conclusions were developed by the authors, who take full responsibility for the contents of the paper. The judging agent codebase, originally written by the authors, was later enhanced with ChatGPT to improve robustness and evaluation reliability, and was reviewed and verified by the authors.

The proposed method is intended for research, benchmarking, and assistive applications. It is not meant to replace human judgment or facilitate academic dishonesty. To promote transparency and reproducibility, all code, evaluations, sampled datasets, and reasoning traces generated by the proposed method are released as open source. We hope that our work will contribute positively to ongoing research in LLM reasoning.

%% file: appendix_sections/theoretical_overview.tex
\section{Theoretical Overview}

\subsection{Error Propagation in Sequential Agents}
Sequential reasoning agents generate solution steps in an ordered manner, where each step $s_t$ depends on prior steps 
$(s_1, s_2, \dots, s_{t-1})$. We can write this as:
\[
s_t = f(s_1, s_2, \dots, s_{t-1}),
\]
where $f$ is the language model. Once a step $s_n$, with 
$1 \leq n \leq t-1$, has been completed, it remains permanent and cannot be revised. As a result, the conditional probability that the next step is correct, $\mathbb{P}(s_t \mid s_{1:t-1})$, reduces as errors accumulate in earlier steps. This causes initial inaccuracies to propagate through the rest of the reasoning trajectory without any opportunity for correction.

\subsection{Iterative Program Refinement and Reflection Memory}

Let $\mathcal{P}$ denote the set of all candidate programs and $I:\mathcal{P}\!\to\!\mathcal{O}$ be the interpreter that produces an output from a program (including errors) $o = I(p)$. At refinement step $t$, the agent has a memory $M_t$ that summarizes past mistakes or errors.

The IIPC refinement operator is then
\begin{align}
p_{t+1}
&= f_{\theta}\!\big(x,\; p_t,\; o_t,\; M_t\big) \\
&= \arg\min_{p \in \mathcal{N}(p_t)}
\Big(
\underbrace{\mathcal{L}\!\big(I(p)\big)}_{\text{execution loss}}
+ \lambda\,\underbrace{\Psi\!\big(p;\,M_t\big)}_{\text{memory penalty}}
\Big).
\end{align}
where $\mathcal{N}(p_t)$ is the complete set of edits around $p_t$, 
$o_t$ is the execution output at step $t$ (programmatic output and errors), 
$\Psi(\cdot;M_t)\!\ge\!0$ penalizes parts of the program that $M_t$ should prevent, $\mathcal{L}\!\big(I(\cdot)\big)$ penalizes the output of the program for undesirable outputs, and $\lambda\!>\!0$ is a trade-off factor between fixing the current error and not repeating past mistakes. Intuitively, the language model attempts to find the next best program so that it produces desirable output and doesn't repeat past mistakes. We do emphasize that this objective is not explicitly optimized, but rather serves as a conceptual abstraction of how execution feedback and reflection memory bias the LLM's refinement behavior.

After developing $p_{t+1}$, the memory updates with the errors of the previous iteration as
\[
M_{t+1}
= M_t \cup \mathsf{Desc}(p_{t},I(p_{t}))
\]
where $\mathsf{Desc}$ extracts failure descriptions (e.g. banned mathematical methods, reflections on minor errors, API misuse, etc) that are then appended to $M_t$.

\subsection{Search and Regret in Refinement}
Let $\mathcal{U}(M_t) \subset \mathcal{P}$ be the union of possible program regions avoided by the memory store of past mistakes. According to the results in table~\ref{tab:iipc-ablation}, where we isolate the effects of individual IIPC components, the reflection memory does improve the accuracy of the IIPC agent. Motivated by these results, we can model the LLM’s refinement behavior under the penalty function $\Psi(\cdot;M_t)$ as follows: (1) it lowers the probability of generating programs from $\mathcal{U}(M_t)$, (2) it does so by at least some multiplicative factor whenever a program is a flawed, and (3) the LLM maintains diversity in its refinements to explore candidates outside $\mathcal{U}(M_t)$. Under these assumptions, we can qualitatively model the LLM's refinement behavior such that if $\Psi(\cdot;M_t)$ assigns a strictly positive penalty on $\mathcal{U}(M_t)$ and zero otherwise, then the expected probability of revisiting blacklisted regions decreases with $t$. As a result, the cumulative revisit regret
\[
R_T = \sum_{t=1}^T \mathbf{1}\{p_t \in \mathcal{U}(M_{t-1})\}
\]
grows sub-proportionally compared to naive refinement. Therefore, the expected per-step revisit rate decreases as memory accumulates. 

Intuitively, this means that each time the LLM encounters an error, the memory store shifts the probability mass away from that type of failure, making it less likely to be chosen again. Over time, the agent spends a smaller fraction of steps repeating past mistakes, while still preserving the ability to explore new refinements within the edit space $\mathcal{N}(p_t)$. In other words, reflection memory turns trial-and-error into trial-and-improvement, where the system learns from its entire history of failures for a given problem.

\subsection{Dual-Branch Separation}
The objective of the dual-branch design is to find an answer $y^*$ that is consistent with the information from both branches.

Let $x$ denote the given problem, $p$ the refined program to solve the problem, and $o$ the output of $p$.

We can first model the reasoning behavior of the two branches in terms of the implicit distributions over candidate answers $y$ as shown:
\begin{itemize}
    \item $d_P(y|x,p,o)$: The distribution over candidate answer $y$ conditioned on $x$, $p$, and $o$ in the execution-guided reasoning branch.
    \item $d_T(y|x)$: The distribution over candidate answer $y$ conditioned on just $x$ in the token-based reasoning branch.
\end{itemize}

In order to reconcile these two potentially conflicting beliefs over the answer space, we introduce an idealized distribution $d^*$ that takes into account information from both branches. We also introduce a weighting factor $\alpha$ that implicitly reflects how much trust should ideally be placed on the execution-guided branch relative to the token-based reasoning branch. The distribution $d^*$ should remain as consistent as possible with both $d_T$ and $d_P$, while also accounting for $\alpha$. We can formalize this notion through a conceptual disagreement-penalized objective:

\begin{align}
d^*
&= \arg\min_d
\Big[
(1-\alpha)\,D_{\mathrm{KL}}\!\big(d \,\|\, d_T\big) \\
&\qquad\quad
+ \alpha\,D_{\mathrm{KL}}\!\big(d \,\|\, d_P\big)
\Big].
\end{align}

where $\alpha \in [0,1]$ and $D_{\mathrm{KL}}$ is the Kullback-Leibler divergence. This objective penalizes solutions that are strongly supported by only one source of reasoning, unless that source is explicitly prioritized by $\alpha$. Intuitively, this means that under $d^*$, answers that contradict token-level logical constraints or execution-based evidence receive a low probability, while answers reinforced by both branches receive a higher probability. It is important to note that while IIPC does not explicitly compute or optimize this objective, the dual-branch design induces behavior that resembles this form of regularization.

Finally, while $d^*$ serves as a conceptual abstraction for the ideal distribution, we ultimately want an answer $y^*$ that is viewed as the most consistent candidate under the reconciled distribution. This can be formalized through the following objective:

\[
y^* \ = \ \arg\max_y d^*(y).
\]

This abstraction shows how dual-branch separation stabilizes reasoning by producing answers that achieve consistency across both token-level and execution-based reasoning.

\subsection{Effects of Model Capacity on Performance}
Our results show IIPC is strongest on higher-capacity models (Gemini 2.0 Flash, Mistral 24B). This can be theoretically understood as an alignment between model capacity and refinement overhead. If the base LLM’s reasoning capacity is insufficient, the iterative correction signal increases cognitive load beyond what the model can process or was trained for, leading to lower accuracy.

%% file: appendix_sections/extended_experiment_details.tex
\section{Additional Experimentation Details}

\paragraph{Reasons for Dataset Selection.}
To capture a range of mathematical complexity, the main experiments evaluate IIPC on the MATH and AIME datasets, while the GSM8K dataset is used for ablation experiments to study the effects of agentic overhead on simple problems. GSM8K~\cite{cobbe2021gsm8k} contains grade school arithmetic problems and serves as a low-difficulty benchmark for an ablation experiment to understand how additional complexity, such as IIPC’s refinement loop and dual-branch design, affects performance on simple tasks. The MATH dataset~\cite{hendrycksmath2021} provides tests for algebraic manipulation, symbolic reasoning, and multi-step logic. Finally, the AIME dataset~\cite{veeraboina2024aime} evaluates on competition-level problems from 1983–2024 that require creativity, deep conceptual understanding, and nontrivial problem decomposition. Evaluating on this range of datasets allows us to analyze how IIPC scales from straightforward arithmetic reasoning to advanced, competition-grade problem solving, and to identify the conditions under which iterative refinement yields the greatest benefit. Ideally, we aimed to evaluate approximately 1,500 problems per dataset to ensure balanced statistical representation. GSM8K only contains 1319 problems in its testing set, all of which were used in the ablation study. For MATH, while we aimed for uniform coverage across all topic–difficulty combinations (35 combinations total), certain combinations contained fewer available problems. This yielded a final sample of 1,483 problems instead of the intended 1,505 (the closest multiple of 35 to 1500). Finally, the AIME dataset contained a fixed set of 933 problems spanning all available years (1983–2024), and was used in its entirety. 

%% file: appendix_sections/additional_experimental_results.tex
\section{Additional Experimental Results for MATH dataset}

\begin{figure*}[t]
  \centering
  \includegraphics[width=\textwidth]{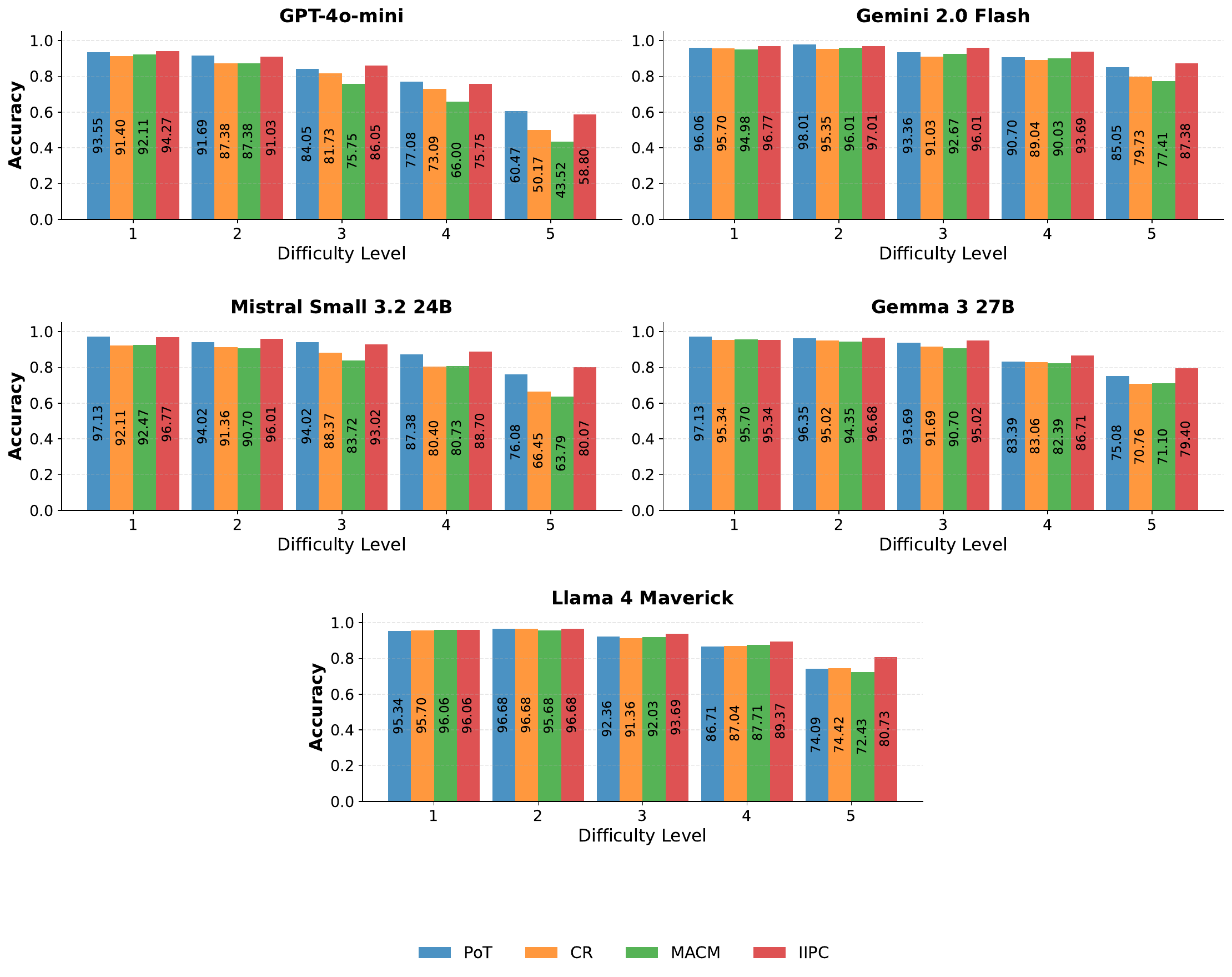}
  \caption{Accuracy by difficulty level for each LLM, comparing PoT, CR, MACM, and IIPC agents on the MATH dataset.}
  \label{fig:math-difficulty-accuracy}
\end{figure*}

\begin{figure*}[t]
  \centering
  \includegraphics[width=\textwidth]{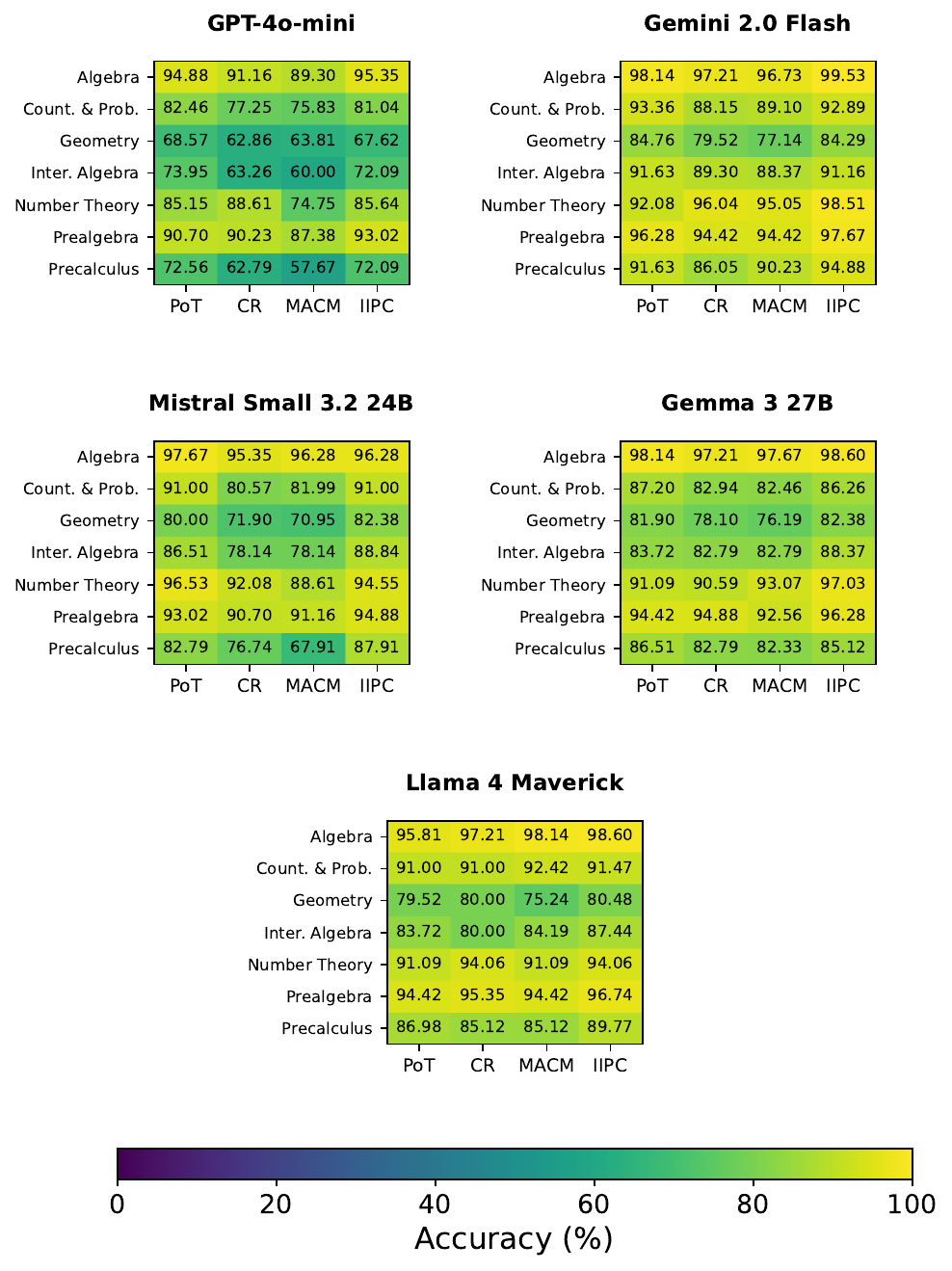}
  \caption{Heatmaps showing accuracy by topic for each LLM, comparing PoT, CR, MACM, and IIPC agents on the MATH dataset.}
  \label{fig:math-topics-accuracy}
\end{figure*}

In our main experiments, we present a fine-grained analysis by problem difficulty and topic only for Llama 4 Maverick. This section extends this same per-difficulty and per-topic comparison to all other language models considered in this paper.

Across difficulty levels (Figure~\ref{fig:math-difficulty-accuracy}), IIPC demonstrates strong performance on Gemini 2.0 Flash, Mistral Small 3.2 24B, Gemma 3 27B, and Llama 4 Maverick, while slightly trailing PoT on GPT-4o mini. On Gemini 2.0 Flash, IIPC maintains high accuracy from 96.77\% at Level~1 to 87.38\% at Level~5, whereas MACM drops from 94.98\% to 77.41\%, a 9.97\% difference at the highest difficulty. Similar trends are observed on Mistral 3.2 24B (IIPC: 96.77→80.07; MACM: 92.47→63.79), Gemma 3 27B (IIPC: 95.34→79.40; MACM: 95.70→71.10), and Llama 4 Maverick (IIPC: 96.06→80.73; MACM: 96.06→72.43), indicating that IIPC better preserves reasoning stability under increasing problem complexity. In contrast, on GPT-4o-mini, PoT maintains an edge over IIPC. We also observe that IIPC maintains the highest accuracy on the most difficult problems on all models except GPT-4o-mini.

The topic-wise analysis (Figure~\ref{fig:math-topics-accuracy}) reveals similar patterns. On Gemini 2.0 Flash, IIPC leads in algebra (99.53\%), number theory (98.51\%), prealgebra (97.67\%), and precalculus (94.88\%). On Gemma 3 27B, IIPC maintains a lead in geometry (82.38\%), intermediate algebra (88.37\%), prealgebra (94.88\%), and precalculus (87.91\%), outperforming other agents by multiple points. On Llama 4 Maverick, IIPC's performance either matches or surpasses that of all other agents across most topics, demonstrating IIPC's ability to fully utilize the reasoning capabilities of high-performing models. On GPT-4o mini, PoT tends to lead IIPC by significant margins.

%% file: appendix_sections/case_studies.tex
\newtcolorbox{SystemPromptBox}{
  enhanced,
  breakable,
  colback=white,
  colframe=black!60,
  boxrule=0.8pt,
  arc=3mm,
  left=4mm,
  right=4mm,
  top=3mm,
  bottom=3mm,
  fontupper=\footnotesize,
}

\section{IIPC Reasoning Examples}

Figure~\ref{fig:gem2flash_prog_ref} is a program refinement generated by Gemini 2.0 Flash when solving an AIME problem. IIPC allows the LLM to reflect on its previous program, identify the incorrect assumption, and propose the fix before developing a new program that produces the correct output.

Figure~\ref{fig:llama4mav_dual} demonstrates the fusion stage of the IIPC agent, where it combines both reasoning branches to produce the final reasoning chain. Here, Llama 4 Maverick makes a mistake in its program by reversing the terms in the formula that calculated the dot product, resulting in a negative value. However, it identifies and applies the correct formula in the token-reasoning branch. In the final reasoning chain, which was conditioned on context from both branches, IIPC allows the LLM to deliberate between both sources and identify the error in the program branch, resulting in a correct final answer.

\clearpage
\begin{figure}[p]
\centering
\begin{minipage}[t][\textheight][t]{\textwidth}
\begin{SystemPromptBox}

\textbf{Problem:}  
A right circular cone has a base with radius 600 and height $200\sqrt{7}$. A fly starts at a point on the surface of the cone whose distance from the vertex of the cone is 125, and crawls along the surface of the cone to a point on the exact opposite side of the cone whose distance from the vertex is $375\sqrt{2}$. Find the least distance that the fly could have crawled.

\vspace{2.0em}

\textbf{Original Program \textcolor{red}{\ding{55}}:}
\begin{verbatim}

python code ...

# The angle difference is pi.
d = math.sqrt(r1**2 + r2**2 - 2 * r1 * r2 * math.cos(math.pi))

python code ...

\end{verbatim}

\vspace{2.0em}

\textbf{Original Program Output \textcolor{red}{\ding{55}}:} 
\begin{verbatim}
The slant height of the cone is: 800.0
The central angle of the sector is: 4.71238898038469
The shortest distance the fly could have crawled is: 
655.3300858899107
655.3300858899107
\end{verbatim}

\vspace{2.0em}

\textbf{Reflection:} 
The previous program calculates the slant height and central angle correctly. However, it incorrectly assumes that "exact opposite side of the cone" translates to an angle difference of $\pi$ in the flattened sector. The correct angle difference should be $\pi$ radians *only if* the central angle of the sector is $2\pi$. In general, the angle difference should be $\pi * (2\pi / \theta)$, where $\theta$ is the central angle of the sector. The program also lacks verification that the final answer is reasonable.

\vspace{2.0em}

\textbf{Refined Program \textcolor{green!70!black}{\ding{51}}:}

\begin{verbatim}
python code ...

# Calculate the angle difference.
delta_theta = math.pi * r / l
print(f"The angle difference is: {delta_theta}")

# Calculate the shortest distance using the law of cosines.
d = math.sqrt(
    r1**2 + r2**2 - 2 * r1 * r2 * math.cos(delta_theta)
)

python code ...

\end{verbatim}

\textbf{Refined Program Output \textcolor{green!70!black}{\ding{51}}:} 
\begin{verbatim}
The slant height of the cone is: 800.0
The central angle of the sector is: 4.71238898038469
The angle difference is: 2.356194490192345
The shortest distance the fly could have crawled is: 625.0
The calculated distance is less than the sum of the distances 
from the vertex, which is expected.
Performing alternative verification...
Approximate shortest distance: 624.9134519586773
\end{verbatim}

\end{SystemPromptBox}
\caption{Example of IIPC's program refinement mechanism using Gemini 2.0 Flash. Only significant changes in the code are displayed to control figure size. The full example text is provided in the GitHub repository.}
\vfill
\label{fig:gem2flash_prog_ref}
\end{minipage}
\end{figure}

\clearpage
\begin{figure}[p]
\centering
\begin{minipage}[t][\textheight][t]{\textwidth}
\begin{SystemPromptBox}

\textbf{Problem:}  
In triangle $ABC,$ $AB = 3$ and $AC = 5.$  Let $O$ be the circumcenter of triangle $ABC.$  Find $\overrightarrow{OA} \cdot \overrightarrow{BC}.$

\vspace{1.5em}

\textbf{Program Branch (Contains Flawed Logic \textcolor{red}{\ding{55}}):}
\begin{verbatim}
import numpy as np

# Define vectors AB and AC
AB = np.array([3, 0])  # Assuming AB lies on the x-axis

# Since we're working with a general triangle, 
# let's consider the angle between AB and AC
# We'll directly use the property that OA . BC = (AC^2 - AB^2)/2

# Calculate OA . BC using the formula derived from the 
# properties of the circumcenter and dot product
AC_length = 5
AB_length = 3
OA_dot_BC = (AC_length**2 - AB_length**2) / 2

print("The dot product of OA and BC is: ", OA_dot_BC)

# Verification
assert np.isclose(
    OA_dot_BC,
    (5**2 - 3**2) / 2
), "Incorrect calculation"
\end{verbatim}

\vspace{1.5em}

\textbf{Original Program Output \textcolor{red}{\ding{55}}:} 
\begin{verbatim}
The dot product of OA and BC is:  8.0
\end{verbatim}

\vspace{1.5em}

\textbf{CoT Branch Text \textcolor{green!70!black}{\ding{51}}:} 

Step 1: To solve the problem, we need to start by understanding the given information and what is being asked. We are dealing with triangle $ABC$, where $AB = 3$ and $AC = 5$. The circumcenter of the triangle is $O$, and we are required to find the dot product of $\overrightarrow{OA}$ and $\overrightarrow{BC}$.

Step 2: First, let's recall that the circumcenter $O$ of a triangle is the point where the perpendicular bisectors of the sides of the triangle intersect. This means $O$ is equidistant from the vertices $A$, $B$, and $C$.

\dots

Step 13: Now, we substitute $AB = 3$ and $AC = 5$ into the expression to find $\overrightarrow{OA} \cdot \overrightarrow{BC}$.

Step 14: $\overrightarrow{OA} \cdot \overrightarrow{BC} = \frac{3^2 - 5^2}{2} = \frac{9 - 25}{2} = \frac{-16}{2} = -8$.

The final answer is: $\boxed{-8}$

\vspace{1.5em}

\textbf{Final Reasoning Chain (Using Both Context Sources) \textcolor{green!70!black}{\ding{51}}:}

The objective of the given problem is to find the dot product of vectors $\overrightarrow{OA}$ and $\overrightarrow{BC}$ in triangle $ABC$, where $AB = 3$, $AC = 5$, and $O$ is the circumcenter of the triangle.

Upon analyzing the context from both the code and the reasoning chain, it is evident that they both lead to the same formula for calculating $\overrightarrow{OA} \cdot \overrightarrow{BC}$, which is $\frac{AC^2 - AB^2}{2}$. The reasoning chain provides a detailed step-by-step derivation of this formula using properties of the circumcenter and dot product.

The code directly implements this formula to calculate $\overrightarrow{OA} \cdot \overrightarrow{BC}$. However, upon closer inspection, it is noticed that the code contains a logical error in its output. The correct calculation should yield a negative result because $AC > AB$, and the formula is $\frac{AB^2 - AC^2}{2}$.

To correct this, we should negate the output of the code or directly use the correct formula. The correct calculation is $\frac{3^2 - 5^2}{2} = \frac{9 - 25}{2} = -8$.

Combining the insights from both the code and the reasoning chain, we arrive at the final answer. The code is genuinely helpful in solving the problem and, with a correction, produces an accurate output.

The final answer is: $\boxed{-8}$

\end{SystemPromptBox}
\caption{Example of IIPC's dual branch mechanism using Llama 4 Maverick. CoT output has been shortened to control image size. The full example text is provided in the GitHub repository.}
\vfill
\label{fig:llama4mav_dual}
\end{minipage}
\end{figure}